\documentclass{llncs}
\usepackage{graphicx}
\usepackage{mathtools}
\usepackage{amsmath}
\usepackage{listings}
\begin{document}

\title{A Survey of Multithreading Image Analysis}
\titlerunning{Hamiltonian Mechanics}  
%


\author{Elham Sagheb Hossein Pour}

\authorrunning{Ivar Ekeland et al.} 
%

%
\institute{Department of Computer Science, University of Wisconsin Milwaukee, USA}

\maketitle              

\begin{abstract}
Digital image analysis has made a big advance in many areas of enterprise applications including medicine, industry, and entertainment by assisting the extraction of semantic information from digital images. However, its large computational complexity has been a trouble to most real-time developments. While image analysis in general has been studied for a log period in computer science community, the use of multithreading strategy as the most efficient improving computational capacity technique has been limited so far. In this survey an attempt is made to explain the current knowledge and so far progresses in incorporating image analysis with multithreading approaches. The present work also provides insights and tendencies for the possible future enhancement of multithreading image analysis. 
\end{abstract}
\section{Introduction}
Image analysis as the intersection of digital signal processing, mathematical models, and machine perception has made a big advance in variety of scientific and commercial applications consisting biomedicine, medicine, and industry by assisting the extraction of semantic information from digital images. The task of image analysis broadly lies from low level image processing such as image sharpening and contrast enhancement, to medium and perhaps high level processing including image segmentation, image registration, object tracking, and image watermarking. A challenge in image analysis applications is to tackle the large computational complexity of almost every image processing task to make such a real-time application. 
In the other side, the multithreading technique has been a very longstanding approach to employ instruction level parallelism which allows having several threads within the context of one single process and it definitely improves the computational capacity by concurrent execution of threads. 
Over the past few years, using multithreading strategies has found significant interest from the variety of scientific researches. This is obviously evident from Figure 1 that shows the journal and conference papers related to the rubric which have been published in Elsevier from 2012 to 2014. 
Using multithreading techniques in the field of digital image analysis can be fulfilled the following objectives:

\begin{itemize} \itemsep.2em

\item [$\bullet$] To facilitate the production of real-time image analysis applications.

\item [$\bullet$] To provide a capability for improving the computational aspects of digital image analysis. 

\item [$\bullet$] To create scalable image analysis applications which can expand and work across multiple clusters.

\item [$\bullet$] To make distributed infrastructures for large scale digital image analysis. 

\end{itemize}

While digital image analysis algorithms and its applications have been studied for a long period, the use of multithreading techniques in image processing has been limited so far. In this work, a survey is performed to review the current knowledge and so far progresses in incorporating digital image analysis with multithreading approaches. With the present paper, I also draw some possible enhancements for multithreading image analysis.

\begin{figure}
  \caption{Number of Elsevier publications on multithreading over the last three years. Results obtained by submitting query “multithreading” from Elsevier website at http://www.sciencedirect.com.}
  \centering
  \includegraphics [height=6.2cm, width=12cm]{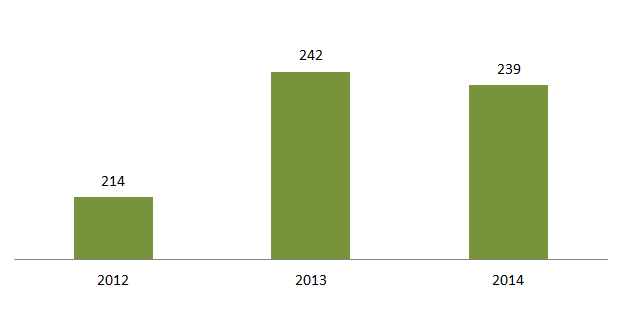}
\end{figure}

The rest of this survey is arranged as follows. In Section 2, a brief introduction for both digital image analysis and multithreading approach is presented. The recent works and advances in the field of multithreading image analysis are reviewed in Section 3. I discuss and conclude the work in Section 4.

\section{An Introduction to Digital Image Analysis and Multithreading Approach}

 While image analysis and its applications have been around for almost 50 years now, multithreading techniques dates back to about 24 years ago. In this section, a brief introduction on these rubrics is presented. 
 
 \subsection{Digital Image Analysis}
 
 Today, we can see digital images everywhere form biomedical as well as astronomical domain to artistic applications. Digital image analysis enables the multimedia technology revolution we are experiencing these days. Some important examples of image processing include image sampling and quantization, image filtering and correlation, coloring, image segmentation, morphological image processing, noise reduction and restoration, feature extraction and object recognition tasks.
All image analysis tasks are divided into three different categories: 1) Low-level, 2) Medium-level, and 3) High-level \cite{lvq1} and \cite{lvq2}. In low-level image analysis techniques, we consider a digital image as an input and produce a digital image as output. It means that the input and output of this level are both digital images. Some examples include noise reduction and contrast enhancement. Using medium-level image analysis, we get an image as an input and provide/extract some information form that. The output of this level is not an image. Image segmentation and object detection such as face detection are some examples. The most difficult part of the image analysis tasks is high-level one in which the input is an image but the output will be knowledge! For instance, an input could be a portrait image from a person’s face, so the output could be whether the person is happy or sad. 

This is a very introductory level of digital image analysis. Readers interested in image analysis algorithms and its applications are referred to \cite{reg01} and \cite{reg02} for image registration,  \cite{seg01}, \cite{seg02}, \cite{seg03}, \cite{seg04}, and \cite{seg05} for image segmentation,  \cite{fr02}, \cite{fr03}, \cite{fr05}, and \cite{fr06} for image forgery detection and image encryption, \cite{t01} and \cite{t02} for 3D surface reconstruction. Further information can be found in \cite{lvq1} and \cite{lvq2}.

\subsection{Multithreading Approach}

A Thread can define as a path of execution through a program. Single threaded application has only one path of execution, while multi-threaded application has two or more path of execution. In such traditional processes, there is only a unique thread of control and a single Program Counter in every process. These processes can perform only one task at a time, and should finish each task before they can start another in the sequence. Using multithreading techniques, the system can support for multiple threads of control for a process at a time. Multithreading offers several advantages as follows \cite{m01}:

\begin{itemize} \itemsep.2em

\item [$\bullet$] Better utilization for employing system resources.

\item [$\bullet$] Improved performance and concurrency. 

\item [$\bullet$] Simultaneous access to multiple applications.

\item [$\bullet$] Task Parallelizations. 

\end{itemize}

Readers interested in multithreading techniques and their implementation can refer to \cite{m02} and \cite{m03} for more information. 

\section{Literature Review}

The following literature review aims to show the current knowledge and so far progress in incorporating multithreading strategies and digital image analysis. 

In 1998, Yu et al. \cite{m04} developed an image convolution algorithm with traditional parallel methods. Their algorithm dramatically reduces computation cost. They discussed two parallel algorithms for performing image convolution, and concluded that using parallel techniques will speed up the image convolution task. However, they mentioned that selecting an algorithm for all application parameters is quite hard. 

In 2003, Penha et al. \cite{m05} performed a comparative study on multithreading image analysis based on shared-variables programming approach. They employed explicit compiler directives from multi-thread support libraries. Their comparison between the implementations has done by considering two kinds of well-known operating systems: Windows and Linux. They have examined both general performance and programmability. The image convolution implementations experiments showed significant performance improvement over sequential ones both in Windows and Linux. The programmability analysis showed that it is simpler for the programmer to develop p-thread based applications rather than another types of thread such as win threads. In general, they showed that using multithreading implementation will improve the general performance, but the implementation itself could not be convenient and easy for developers.

In 2009, Lin et al. \cite{m06} carried out a multithreading strategy as parallel method to perform the PDE-based image registration. They implemented deformable image registration and examine it on a dual core personal computer. For implantation, they used OpenMP as an API which can support multi-processing programming in C++. Their experiments demonstrated that the method was able to produce the large size parallel image registration, reduce the computational complexity, and save nearly a half computing time, however the implementation part was not easy enough. 

In 2010, GG Slabaugh et al. \cite{m07} presented the entire pipeline of using OpenMP for doing image processing tasks such as image morphology, image filtering, and normalization. They summarized the general capabilities of OpenMP and showed that Signal and image processing programmers can benefit dramatically from multithreading techniques provided through OpenMP, by modifying a single-threaded code to operate parallelism.
In 2013, Kika et al. \cite{m08} used Java programming to deploy digital image analysis tasks in single-core and multi-core CPU. They mentioned that the Java programming language is very appropriate to build and develop image analysis applications due to its features and the free and open source packages that it offers for this purpose. Their experimental results showed that such a multithreading method will definitely improve the performance of image analysis algorithms either in single-core or multi-core CPU platforms, however the improvements are different.  In single-core, the best results is achieved by the combination of small image size and less complex algorithm, while in multi-core CPU the combination of small image size and more complex algorithm improves the performance. They also concluded that the multithreading programming can improve the performance on multi-core CPU whenever complex image processing algorithms is applied.

In 2015, Smistad et al. \cite{m09} developed an open-source efficient medical image analysis and visualization framework namely FAST (FrAmework for heterogeneouS medical image compuTing and visualization) using multithreading techniques. The code examples along with the evaluations have demonstrated that the framework is easy to use and performs better than existing frameworks, such as ITK and VTK.

 \section{Discussion, Conclusion, and Possible Extension}

In this work, a survey is done to review the recent works in multithreading image analysis. After a brief introduction to digital image analysis and multithreading areas, five recent works multithreading image processing are reviewed. In almost every work, we can see that using multithreading strategies will improve the general performance of digital image analysis tasks such as image convolution, image filtering, and morphology either in single-core or multi-core CPU. The implementation multithreading techniques from scratch or using pre built open source library is still remain difficult. In general, multithreading approaches will improve the performance and time efficiency of the image analysis tasks and allows resource utilization. Difficulties in implementation, debugging, and managing concurrency are among some multithreading image analysis disadvantages. 
Multithreading image analysis using Big Data infrastructures would systematically provide better performance and reliability to process Big Data images, such as very high resolution satellite images. Employing SaaS (Software as a Service) architecture would also provide application-to-application interaction for such systems. These could be considered as possible extensions for multithreading image analysis.

%
%

\end{document}